\documentclass[10pt]{article}

\usepackage[utf8]{inputenc}
\usepackage[T1]{fontenc}
\usepackage{lmodern}
\usepackage[a4paper, margin=1in]{geometry}
\usepackage{microtype}
\usepackage{hyperref}
\hypersetup{colorlinks=true, linkcolor=black, citecolor=black, urlcolor=black}
\usepackage{url}
\usepackage{booktabs}
\usepackage{amsmath}
\usepackage{amssymb}
\usepackage{amsfonts}
\usepackage{nicefrac}
\usepackage{graphicx}
\usepackage{xcolor}
\usepackage{titlesec}
\usepackage[numbers,square,sort&compress]{natbib}
\usepackage{enumitem}
\usepackage{caption}
\usepackage{xspace}
\usepackage{parskip}
\usepackage{tabularx}
\newcolumntype{Y}{>{\raggedright\arraybackslash}X}

\usepackage{newunicodechar}
\newunicodechar{Δ}{\ensuremath{\Delta}}

\titleformat*{\section}{\Large\bfseries}
\titleformat*{\subsection}{\large\bfseries}
\titleformat*{\subsubsection}{\normalsize\bfseries}

\title{\Large\bfseries Procedural-Skill SFT Across Capacity Tiers:\\
  A W-Shaped Pre-SFT Trajectory and Regime-Asymmetric Mechanism\\
  on 0.8B--4B Qwen3.5 Models}

\author{Igor Strozzi\thanks{\texttt{istrozzi@matematica.ufrj.br}}\\
  Applied Mathematics Department\\
  Federal University of Rio de Janeiro}

\date{May 6, 2026}

\newcommand{\BL}{\textsc{baseline}\xspace}
\newcommand{\CU}{\textsc{curated}\xspace}

\newcommand{\Qwent}{Qwen3.5-2B\xspace}
\newcommand{\Qwenf}{Qwen3.5-4B\xspace}

\begin{document}
\maketitle

\begin{abstract}
We measure procedural-skill SFT contribution across three Qwen3.5 dense scales (0.8B, 2B, 4B) on a 200-task / 40-skill holdout, with Claude Haiku 4.5 as a frontier reference. The corpus is 353 rows of (task + procedural-skill block, Opus chain-of-thought, judge-pass) demonstrations.

\textbf{Main finding.} Under matched-path LLM-only scoring, the SFT-attributable procedural-Δ lift is roughly uniform across the three sizes: $+0.070$ / $+0.040$ / $+0.075$ at 0.8B / 2B / 4B. Variation in post-SFT Δ across sizes ($-0.005$, $+0.100$, $+0.065$) is dominated by a W-shaped pre-SFT base trajectory ($-0.075$, $+0.060$, $-0.010$, with frontier Haiku-4-5 at $+0.030$): the 5-step procedure hurts very small (0.8B) and competent-but-not-frontier (4B) bases, helps an intermediate (2B) base, and helps a frontier base modestly. SFT works hardest in absolute terms where the base struggles with the procedure --- a regime-asymmetric pattern with a falsifiable prediction at 8B/14B.

\textbf{Methodological observations.} (i) A bench format-compliance artifact: 83.5\% of the holdout uses a deterministic \texttt{ANSWER}-line extractor that under-counts free-form-prose conclusions. Our LLM-only re-judge closes the gap and reveals the deterministic judge had been systematically biased \emph{against} the curated condition. (ii) A negative-iteration sequence at 0.8B: three well-formed recipe variants (varying corpus composition, partial fine-tuning, skill-block stripping) cluster post-SFT \CU{} pass-rate within a 2\,pp band, constraining the absolute-pass-rate ceiling to base capacity rather than recipe.

\textbf{Cross-family judge validation.} GPT-5.4 via OpenRouter on all 7 configurations (2800 paired episodes) agrees on the direction of every per-student finding: Cohen's $\kappa \geq 0.754$, agreement $\geq 93.25\%$, max headline Δ shift $\leq 0.035$\,pp.

Two earlier framings --- ``format-only learning at 0.8B'' and ``SFT contribution shrinks at 4B'' --- were path-mismatch artifacts; this paper supersedes both (Appendix~\ref{sec:appendix-path}). Single-seed evaluation; threats itemised in \S\ref{sec:threats}.
\end{abstract}

\section{Introduction}

A common compositional-skills hypothesis holds that LLM capabilities can be decomposed into reusable skills \citep{yu2024skillmix,arora2023theory}, and that explicit procedural instructions can elicit better task performance from smaller students than implicit-skill priming alone \citep{li2026skillsbench}. A natural follow-up: does demonstrating the explicit procedure during fine-tuning teach a smaller student to \emph{use} such procedures more effectively at inference time, or merely to imitate the response shape?

We answer this empirically on a self-contained pipeline: a hand-curated 40-skill procedural catalog, an Opus-synthesized 200-task holdout, a 353-row Opus-traced curated SFT corpus, three Qwen3.5 LoRA students (0.8B/2B/4B), and a matched-path evaluation protocol with both deterministic and LLM-judge scoring. Throughout, Opus 4.7 was used for skill rewriting, task synthesis, baseline trace generation, and judge-of-record scoring; this judge-overlap is a structural threat to validity we address quantitatively via cross-family validation (\S\ref{sec:cross-family}) and itemise explicitly (\S\ref{sec:threats}).

\textbf{Empirical contributions.}
\begin{enumerate}[leftmargin=*]
  \item \textbf{Cross-scale SFT contribution under matched-path scoring.} The SFT-attributable procedural-Δ lift is roughly uniform at $+0.070$ / $+0.040$ / $+0.075$ across 0.8B / 2B / 4B --- a 4\,pp band across an order of magnitude in base capacity (\S\ref{sec:cross-scale}).
  \item \textbf{W-shaped pre-SFT base trajectory and regime-asymmetric SFT mechanism.} Pre-SFT Δ is non-monotone with two negative pockets (0.8B and 4B). SFT works hardest where the base struggles with the procedure, and less where the base is already in a procedure-friendly regime --- a falsifiable prediction at 8B/14B (\S\ref{sec:mechanism}).
  \item \textbf{v2.0 ties the frontier reference at the bench's effective ceiling.} v2.0 \CU{} $= 0.985$ matches Haiku-4-5 \CU{} $= 0.985$ (197/200 identical pass count); pre-SFT 4B \CU{} $= 0.840$ rules out the deflationary alternative that base scaling alone closes the gap (\S\ref{sec:haiku}).
\end{enumerate}

\textbf{Methodological contributions.}
\begin{enumerate}[leftmargin=*, resume]
  \item \textbf{Bench format-compliance artifact.} The deterministic ANSWER-line extractor used by 83.5\% of the bench under-counts free-form-prose conclusions, biasing similar SFT-on-procedural-demonstrations claims absent a format-tolerant base-model control. Our LLM-only re-judge reveals the bias was systematically against the curated condition (\S\ref{sec:format-artifact}).
  \item \textbf{Negative-result iteration sequence at 0.8B.} Of five recipe variants tested (varying corpus composition, partial fine-tuning, and skill-block stripping), the three well-formed variants cluster post-SFT \CU{} within 2\,pp; the other two are anomalous failure modes diagnosed in \S\ref{sec:0.8b-iteration}. The 0.8B absolute-pass-rate ceiling is base-capacity-bound, not recipe-bound.
\end{enumerate}

We release the pipeline, training/eval scripts, recipes, and full episode-level logs.

\section{Pipeline and Bench}

\subsection{Skill catalog, procedural rewrite, and task synthesis}
We hand-curate 40 declarative skills across categories (logical fallacies, deductive forms, cognitive biases, interpretive patterns, spatial reasoning), seeded from Skill-Mix Table 5/6 \citep{yu2024skillmix}. Opus 4.7 (1M context) rewrites each declarative entry into a procedural form following SkillsBench \citep{li2026skillsbench}: ordered procedure steps, when-to-use criteria, constraints, and a positive/negative worked example. A task synthesizer prompts Opus to produce 15 tasks per skill; the 600 generated tasks are split into 400 train / 200 eval, disjoint by task-uid with a 1:1 task-to-skill map. Tasks carry one of four query types: \texttt{YES\_NO} (61\%), \texttt{FREE\_FORM} (16.5\%), \texttt{SINGLE\_WORD} (11.5\%), \texttt{RANKING} (11\%). Each task is generated knowing its target skill; the bench therefore measures procedural application on procedure-aligned tasks, not skill-routing or compositional skill use (\S\ref{sec:threats}).

\subsection{SFT corpus and evaluation protocol}
Each training task is traced by Opus under a \textsc{solver} system prompt with the corresponding procedural skill block injected. Episodes that pass the LLM-judge \citep{zheng2023judging} are filtered into a 353-row curated-only chat-format corpus: \texttt{(system: SOLVER + skill block, user: task, assistant: CoT + ANSWER line)}.

For evaluation, we generate one student response per \texttt{(task, condition)} pair on the 200-task holdout. Two conditions: \BL{} (\textsc{solver} prompt without skill block) and \CU{} (\textsc{solver} prompt with the matching procedural skill block). Decoding: greedy ($T{=}0$), \texttt{repetition\_penalty}\,$=1.05$, \texttt{max\_new\_tokens} 1024 post-SFT / 2048 pre-SFT.

Judge dispatch has two paths. \textit{Deterministic-mixed} (the bench's native dispatch): deterministic \texttt{ANSWER}-line extraction for \texttt{YES\_NO}/\texttt{SINGLE\_WORD}/\texttt{RANKING}, Opus 4.7 LLM-judge for \texttt{FREE\_FORM}. \textit{LLM-only}: Opus 4.7 LLM-judge for every task type, used in \S\ref{sec:format-artifact} to bypass the format-compliance gate. We report pass-rate (binary verdict) and $\Delta = \mathrm{pass\_rate}(\CU) - \mathrm{pass\_rate}(\BL)$. All numbers are single-seed.

\subsection{Training}
LoRA \citep{hu2022lora} via TRL\,$\geq$\,0.18 \citep{trl} and PEFT\,$\geq$\,0.13. Adapters target all-linear modules. The chat template is patched with \texttt{\{\% generation \%\}\dots\{\% endgeneration \%\}} markers to enable assistant-only loss masking. For partial-FT variants (v1.7, v1.8) we freeze all but the top-$N$ transformer layers + \texttt{lm\_head} + final \texttt{norm}, using bitsandbytes 8-bit Adam \citep{dettmers2023qlora}. Base models are the Qwen3.5 small dense series \citep{qwen35release}, which builds on the Qwen3 architecture \citep{qwen3report}.

\section{Cross-Scale SFT Contribution and the W-Shape}\label{sec:cross-scale}

We measure the SFT contribution at each model scale by comparing each post-SFT student against its matched-path pre-SFT control under LLM-only scoring. Table~\ref{tab:sft-contribution} reports the contribution decomposition; Table~\ref{tab:llm-only} gives the underlying pass rates for all seven configurations.

\begin{table}[h]
\centering
\small
\begin{tabular}{@{}l rrr rrr@{}}
\toprule
& \multicolumn{3}{c}{\textbf{Pre-SFT base (LLM-only)}} & \multicolumn{3}{c}{\textbf{SFT contribution}} \\
\cmidrule(lr){2-4} \cmidrule(lr){5-7}
& BL & CU & $\Delta$ & $\Delta$BL & $\Delta$CU & $\Delta$-lift \\
\midrule
0.8B (v1)   & 0.665 & 0.590 & $-0.075$ & $+0.045$ & $+0.115$ & $\boldsymbol{+0.070}$ \\
2B (v1.9)   & 0.755 & 0.815 & $+0.060$ & $+0.060$ & $+0.100$ & $\boldsymbol{+0.040}$ \\
4B (v2.0)   & 0.850 & 0.840 & $-0.010$ & $+0.070$ & $+0.145$ & $\boldsymbol{+0.075}$ \\
\bottomrule
\end{tabular}
\caption{SFT contribution across all three model sizes under matched HF + LLM-only scoring. The Δ-lift attributable to SFT is in a tight 4\,pp band ($+0.040$ to $+0.075$); the absolute \CU{} contribution is in a similar band ($+0.100$ to $+0.145$). Variation in post-SFT Δ across sizes is dominated by the pre-SFT base trajectory, which is W-shaped (negative at 0.8B and 4B, positive at 2B and frontier Haiku-4-5; see Table~\ref{tab:llm-only}).}
\label{tab:sft-contribution}
\end{table}

\begin{table}[h]
\centering
\small
\begin{tabular}{@{}l r r r c c@{}}
\toprule
& \textbf{BL} & \textbf{CU} & \textbf{$\Delta$} & \textbf{McNemar $p$} & \textbf{bootstrap 95\% CI on $\Delta$} \\
\midrule
\textit{Pre-SFT base controls} \\
\textbf{pre-SFT 0.8B (HF)} & \textbf{0.665} & \textbf{0.590} & $\boldsymbol{-0.075}$ & --- & --- \\
pre-SFT 2B (HF)            & 0.755 & 0.815 & $+0.060$ & --- & --- \\
\textbf{pre-SFT 4B (HF)}   & \textbf{0.850} & \textbf{0.840} & $\boldsymbol{-0.010}$ & --- & --- \\
pre-SFT Haiku-4-5 (ref)    & 0.955 & 0.985 & $+0.030$ & --- & --- \\
\addlinespace
\textit{Post-SFT students} \\
v1   (0.8B+LoRA)   & 0.710 & 0.705 & $-0.005$ & --- & --- \\
v1.9 (2B+LoRA)     & 0.815 & 0.915 & $+0.100$ & $0.0027$ & $[+0.040,\ +0.160]$ \\
v2.0 (4B+LoRA)     & 0.920 & 0.985 & $+0.065$ & $0.00098$ & $[+0.030,\ +0.105]$ \\
\bottomrule
\end{tabular}
\caption{Pass rates for all seven configurations under matched HF + LLM-only scoring. Haiku-4-5 was evaluated via the Anthropic API; all Qwen3.5 configurations were evaluated via HF transformers locally. Three load-bearing observations: (i) v2.0 \CU{} ties the Haiku reference at $0.985$ (197/200 identical pass count); (ii) pre-SFT base Δ is W-shaped --- $-0.075, +0.060, -0.010, +0.030$ --- with negative pockets at 0.8B and 4B; (iii) the SFT contribution to Δ-lift is roughly uniform across all three sizes (Table~\ref{tab:sft-contribution}).}
\label{tab:llm-only}
\end{table}

\begin{figure}[h]
\centering
\includegraphics[width=\linewidth]{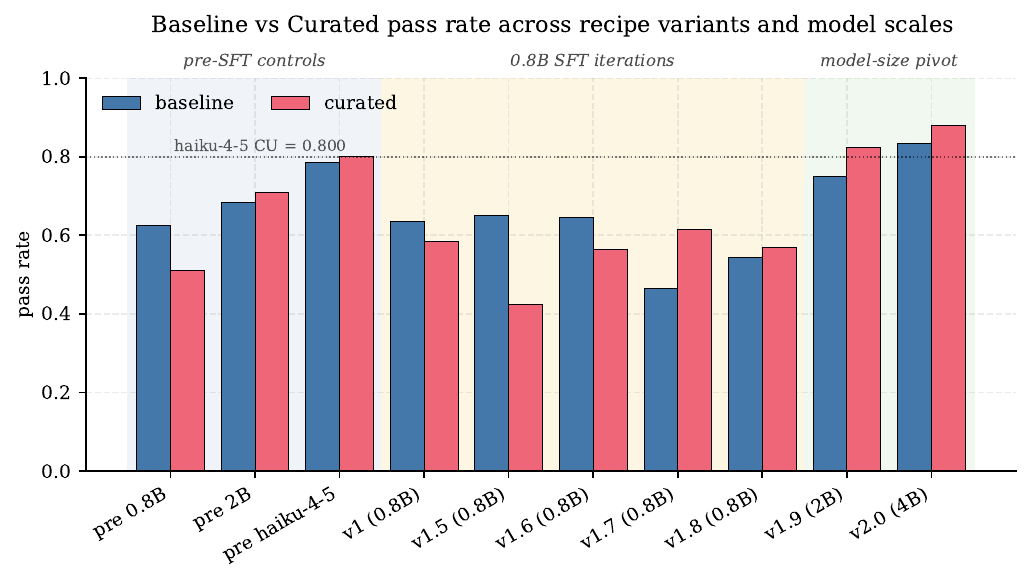}
\caption{Baseline (\BL) and curated (\CU) pass rates across all eleven evaluated configurations: pre-SFT controls, the five 0.8B SFT iterations (\S\ref{sec:0.8b-iteration}), and the model-size pivot to 2B and 4B.}
\label{fig:progression}
\end{figure}

\subsection{The SFT contribution is roughly capacity-invariant in differential terms}\label{sec:uniformity}
Across the three model sizes under matched-path LLM-only scoring, the SFT-attributable Δ-lift sits in a 4\,pp band: $+0.070$ at 0.8B, $+0.040$ at 2B, $+0.075$ at 4B. The absolute \CU{} contribution is in a similar band: $+0.115$ at 0.8B, $+0.100$ at 2B, $+0.145$ at 4B. Per-model McNemar tests under LLM-only judging confirm v1.9 ($\chi^2=9.03$, $p=0.0027$, CI $[+0.040, +0.160]$) and v2.0 (exact, $p=0.00098$, CI $[+0.030, +0.105]$) each have a positive Δ with bootstrap CI excluding zero (Table~\ref{tab:stats-per-model}). The 0.8B SFT Δ-lift is judge-invariant within 1\,pp: $+0.065$ under deterministic-mixed judging, $+0.070$ under LLM-only (\S\ref{sec:format-artifact} explains the per-judge gap).

The differential component is what distinguishes ``the model uses the injected procedure'' from ``the model just imitates the response shape.'' Pure format-learning would lift \BL{} and \CU{} equally and leave Δ untouched, which we do not see at any of the three scales --- Δ-lifts are $+0.070$, $+0.040$, $+0.075$. We resist a quantitative ``half format, half procedure'' decomposition: the format and procedure components are not independent (producing the procedural-step shape implicitly enforces a checking discipline that may itself be the procedural lift), and the underlying contributions cannot be cleanly separated by per-condition pass-rates alone. The differential Δ-lift is evidence that a procedure-vs-no-procedure component of the gain exists; it is not a quantitative apportionment.

\subsection{W-shaped pre-SFT base trajectory}\label{sec:wshape}
The pre-SFT base trajectory of procedural responsiveness is non-monotone with two negative pockets:
\begin{itemize}[leftmargin=*, nosep]
  \item \textbf{pre-SFT 0.8B (HF, LLM-only): $\Delta = -0.075$ --- deepest trough}
  \item pre-SFT 2B (HF, LLM-only): $\Delta = +0.060$
  \item \textbf{pre-SFT 4B (HF, LLM-only): $\Delta = -0.010$ --- shallow trough}
  \item pre-SFT Haiku-4-5 (LLM-only): $\Delta = +0.030$
\end{itemize}
Both 0.8B and 4B are in regimes where the procedure injection actively hurts the base; 2B and Haiku are in regimes where it helps. The two troughs likely reflect different mechanisms:
\begin{itemize}[leftmargin=*, nosep]
  \item \emph{0.8B (deep trough):} the base lacks the capacity to follow the 5-step procedure correctly. Errors compound at every step; the procedure structure adds sub-inference error opportunities the base cannot recover from.
  \item \emph{4B (shallow trough):} the base is competent enough to solve baseline tasks directly. The procedure adds overhead without adding signal --- curated responses sometimes follow the procedure into less-defensible conclusions on tasks the base would have one-shotted.
\end{itemize}
Frontier-class Haiku partially recovers by having enough native capacity to use the procedure as intended without being captive to it. 2B sits in a sweet spot where the procedure scaffolds checking the model would otherwise skip.

\subsection{Regime-asymmetric SFT mechanism}\label{sec:mechanism}
The roughly capacity-invariant Δ-lift magnitude is doing visibly different work at different scales:
\begin{itemize}[leftmargin=*, nosep]
  \item \textbf{0.8B} (pre-SFT $\Delta -0.075$, post-SFT $\Delta -0.005$): SFT lifts by $+0.070$, rescuing a deeply procedure-incompetent regime to procedure-neutral. SFT teaches the model to track the procedure without the compounding errors.
  \item \textbf{2B} (pre-SFT $\Delta +0.060$, post-SFT $\Delta +0.100$): SFT lifts by $+0.040$, marginally improving an already-procedure-friendly regime. The base is in the sweet spot; SFT adds discipline but has less headroom.
  \item \textbf{4B} (pre-SFT $\Delta -0.010$, post-SFT $\Delta +0.065$): SFT lifts by $+0.075$, flipping a procedure-overhead regime into procedure-helps. SFT teaches the model to use the procedure productively rather than treat it as friction.
\end{itemize}
The pattern: SFT works hardest (in absolute Δ-lift terms) where the base struggles with the procedure, and less where the base is already in a procedure-friendly regime. The ``uniform'' framing is fair as a magnitude bound; the underlying mechanism is regime-asymmetric.

\paragraph{Falsifiable prediction at 8B/14B.} If the mechanism explanation is correct, at 8B/14B the procedure should hurt the base less than at 4B (since the base can use it natively, like Haiku), and SFT's Δ-lift should shrink toward 2B-ish levels. If pre-SFT Δ recovers toward Haiku-class positive Δ as base capacity grows, and SFT's Δ-lift correspondingly shrinks in the procedure-friendly regimes, the W-shape and regime-asymmetric story extend. If pre-SFT Δ stays negative or SFT's Δ-lift stays uniform, the mechanism explanation needs revision.

\subsection{Per-skill evidence supports the mechanism}\label{sec:per-skill}
\begin{figure}[h]
\centering
\includegraphics[width=0.85\linewidth]{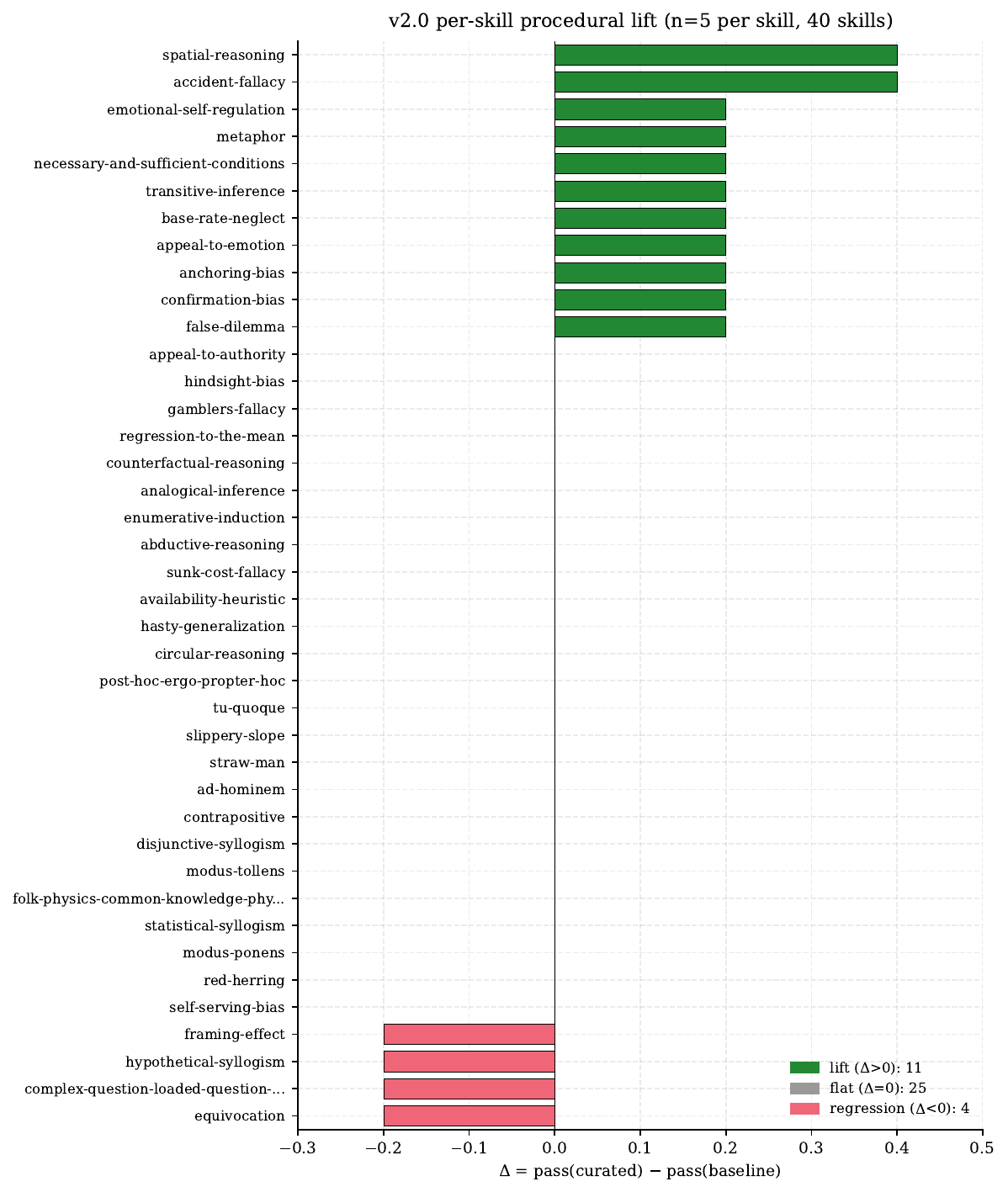}
\caption{Per-skill procedural-Δ at v2.0 ($n{=}5$ per skill, 40 skills). 11 skills lift on \CU; 25 are flat (mostly because \BL{} is already at ceiling); 4 regress. The lift cluster concentrates on interpretive and spatial-social skills; the regression cluster on deductive-logic skills where the model has ceiling competence on \BL.}
\label{fig:per-skill}
\end{figure}

The regime-asymmetric pattern recurs at the skill level. Figure~\ref{fig:per-skill} shows the v2.0 procedural-Δ broken down by skill (deterministic-judge counts). Eleven skills lift on \CU{} (concentrated on interpretive and spatial-social skills); 25 are flat (most because \BL{} is at ceiling and \CU{} has nowhere to go); 4 regress under deterministic judging (deductive-logic skills where the base has ceiling competence, mirroring the 4B pre-SFT trough at the skill level). Under LLM-only judging the regression cluster shrinks to one skill.

\begin{figure}[h]
\centering
\includegraphics[width=0.78\linewidth]{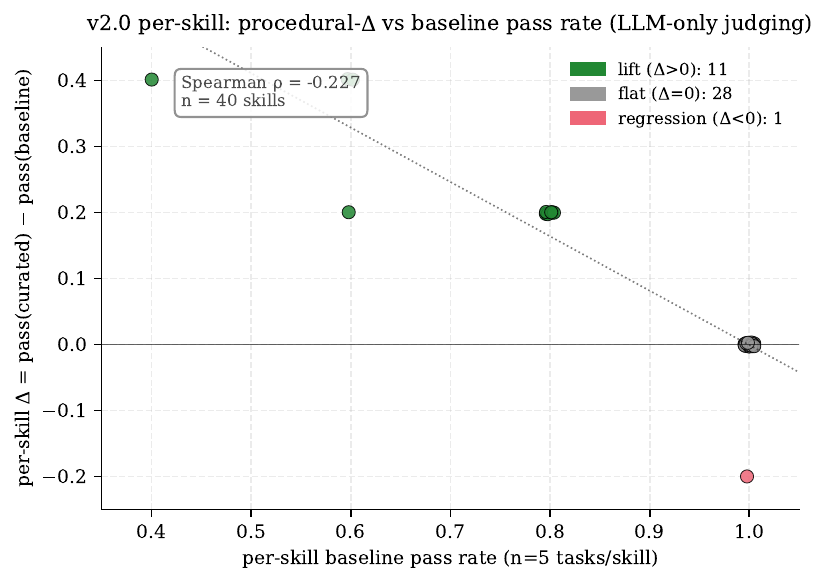}
\caption{Per-skill procedural-Δ at v2.0 versus per-skill baseline pass rate (LLM-only judging, $n{=}5$ tasks per skill, 40 skills). Spearman $\rho = -0.227$. Skills where the base is already at ceiling on baseline cannot benefit from procedure injection (flat cluster); skills with lower baseline have the most Δ headroom (lift cluster). The negative slope is qualitative evidence that SFT's distinctive value is in scaffolding skills the base struggles with, not in lifting already-saturated skills.}
\label{fig:per-skill-bl-vs-delta}
\end{figure}

Figure~\ref{fig:per-skill-bl-vs-delta} plots per-skill Δ against per-skill \BL{} at v2.0 under LLM-only judging. Spearman $\rho = -0.227$ across $n{=}40$ skills (two-sided $p \approx 0.15$, underpowered: 29/40 skills are tied at \BL{}\,$= 1.0$, collapsing to the same rank). On the non-ceiling subset (skills with $\BL < 1.0$, $n{=}11$), the correlation strengthens to $\rho = -0.455$ ($p \approx 0.13$); the negative slope is qualitatively consistent in both subsets but neither reaches conventional significance at this $n$. The lift cluster has mean baseline $0.71$; the flat cluster has mean baseline $1.00$. We treat this as qualitative evidence rather than a load-bearing statistical claim: the procedure helps where the base struggles and is silent where the base is at ceiling --- the per-skill miniature of the cross-scale regime-asymmetric finding.

\subsection{v2.0 ties Haiku at the bench's effective ceiling}\label{sec:haiku}
Under fair scoring, v2.0 \CU{} $= 0.985$ matches the Haiku-4-5 reference at \CU{} $= 0.985$ (197/200 identical pass count). With pre-SFT 4B measured at \CU{} $= 0.840$ under fair judging, the v2.0$=$Haiku tie is attributable to SFT plus base scaling: the 4B base does \emph{not} already match Haiku before SFT (SFT contributed $+0.145$ \CU{} on top of base 4B). An earlier framing where ``2B+SFT crosses Haiku'' under the deterministic-mixed dispatch ($0.825$ vs $0.800$) was an artifact of the format-compliance gate (\S\ref{sec:format-artifact}): Haiku's free-form prose was disproportionately rejected by the \texttt{ANSWER}-line extractor, and under LLM-only judging Haiku's \CU{} rises by $+0.185$ to the ceiling --- more than any other student. Under LLM-only judging, v1.9 \CU{} ($0.915$) does \emph{not} cross Haiku; only v2.0 does.

We resist a sharp compute-efficiency claim relative to Haiku: its parameter count is unpublished (community estimates $10^{10}$--$3{\times}10^{10}$ parameters, no anchor); Opus 4.7 wrote the corpus content, generated the SFT traces, and judged both students --- a structurally biased judging path (\S\ref{sec:cross-family} partially closes this); and the SFT student is in-distribution for the bench while Haiku is asked cold. The honest summary: under fair judging, v2.0 (4B+SFT) and Haiku tie at the bench's curated ceiling on tasks of this construction style.

\subsection{The bench saturation diagnosis splits in two}\label{sec:saturation}
The v1.9$\to$v2.0 post-SFT Δ shrinkage ($+0.100 \to +0.065$ under LLM-only) is directionally consistent with bench saturation but not statistically distinguishable from sampling noise on the v1.9-vs-v2.0 adjacent-pair test: one-sided bootstrap $p = 0.191$ under LLM-only judging ($p = 0.225$ under deterministic-mixed), 95\% CI on $\Delta_{\mathrm{v1.9}} - \Delta_{\mathrm{v2.0}}$ is $[-0.040, +0.105]$, which includes zero. Saturation has two components, and they diverge:
\begin{itemize}[leftmargin=*, nosep]
  \item \emph{Absolute pass rates:} saturated. v2.0 \CU{} $= 0.985$ ties Haiku at the bench's effective ceiling.
  \item \emph{SFT mechanism:} not saturated. Δ-lift is roughly uniform across all three sizes ($+0.040$ to $+0.075$), and grows from 2B's $+0.040$ to 4B's $+0.075$ rather than shrinking.
\end{itemize}
What changes across the post-SFT trajectory is the pre-SFT starting point (W-shape), not the SFT mechanism's strength. An earlier version of this paper conflated absolute saturation with mechanism saturation under one label; only the absolute reading survives.

\section{Format-Compliance Artifact and LLM-Only Remediation}\label{sec:format-artifact}

\subsection{The artifact}
83.5\% of the bench (167/200 tasks) uses a deterministic \texttt{ANSWER}-line extractor for scoring. Base \Qwenf{} reasons correctly on bench tasks (spot-checks confirm step-by-step reasoning that reaches the correct answer) but does not consistently end with \texttt{ANSWER: <X>}, so the deterministic dispatch returns failure. The bench's $0/200$ for pre-SFT 4B under deterministic-mixed scoring should be read as ``format-compliance baseline $\approx 0$,'' not ``reasoning $\approx 0$.''

This generalises: any post-SFT lift measured against a format-strict deterministic judge conflates three contributions --- format learning (the model emits the required line), format-conditional reasoning gating (the model reasons more reliably when constrained to structured output), and procedural application (the model uses the injected skill). Without a base-model control evaluated under format-tolerant scoring, these are not separable. Plausible remedies: (a) a stricter base-model evaluation prompt that reliably hits the format target, (b) a format-tolerant judge for base controls (e.g.\ extracting the answer with an LLM call rather than a regex), (c) restricting evaluation to \texttt{FREE\_FORM} tasks. We apply remedy (b).

\subsection{LLM-only re-judge: closing the format gap}\label{sec:llm-rejudge}
We re-route every episode through the LLM judge regardless of task type. For the post-SFT students and Haiku we re-ran the LLM judge over existing model responses (no regeneration); for the pre-SFT base models we ran fresh HF-transformers evaluations with \texttt{--force-llm-judge}. Numbers are in Table~\ref{tab:llm-only}; Table~\ref{tab:rejudge-deltas} shows the score shifts from deterministic-mixed.

\begin{table}[h]
\centering
\small
\begin{tabular}{@{}l r r r@{}}
\toprule
& $\Delta$BL & $\Delta$CU & $\Delta\Delta$ \\
\midrule
v1   (0.8B+LoRA)   & $+0.075$ & $+0.120$ & $+0.045$ \\
v1.9 (2B+LoRA)     & $+0.065$ & $+0.090$ & $+0.025$ \\
v2.0 (4B+LoRA)     & $+0.085$ & $+0.105$ & $+0.020$ \\
Haiku-4-5 (ref)    & $+0.170$ & $+0.185$ & $+0.015$ \\
\bottomrule
\end{tabular}
\caption{Score shifts from deterministic-mixed to LLM-only re-judge. All four students gain on both conditions; Haiku gains the most. For all three SFT'd students, $\Delta$CU $>$ $\Delta$BL by 2--4.5\,pp, meaning the deterministic judge was systematically biased \emph{against} the curated condition --- procedure-following responses bury conclusions in narration more often than direct-thinking responses do.}
\label{tab:rejudge-deltas}
\end{table}

\paragraph{All pass rates rise; Haiku rises the most.} The deterministic-mixed judge under-counted passes universally. Even the SFT'd students, trained to produce \texttt{ANSWER} format, occasionally emitted alternative phrasings (``Final Answer: X,'' ``I conclude X,'' or stating the answer in prose) that the regex missed. Haiku, never trained on the bench's response format, was penalised the most --- $+17$\,pp on \BL{} and $+18.5$\,pp on \CU{}. This is the cleanest direct evidence that the format-compliance gate biased the bench against students whose response style does not match the SFT distribution.

\paragraph{The deterministic judge was biased against the curated condition.} For all three SFT'd students, $\Delta$CU exceeds $\Delta$BL by 2--4.5\,pp under re-judge (Table~\ref{tab:rejudge-deltas}). When the procedure block is present, the model occasionally embeds its conclusion inside procedural narration (``\dots therefore the conclusion follows: X'') rather than ending with a literal \texttt{ANSWER: X} line. The deterministic judge under-counted curated wins systematically, under-stating the SFT contribution. Under LLM-only judging, v2.0's $p{=}0.00098$ comfortably survives Bonferroni correction across the two per-model tests; the deterministic-mixed $p{=}0.049$ would not.

\paragraph{0.8B path-mismatch implication.} The original pre-SFT 0.8B baseline was run via Ollama on a different code path than the v1 post-SFT evaluation, producing artificially low baseline numbers and an apparent ``Δ-flip'' that earlier versions of this paper read as ``format-only learning at 0.8B.'' Matched-path HF + LLM-only scoring resolves the mismatch and reveals the SFT Δ-lift at 0.8B is $+0.070$ (LLM-only) or $+0.065$ (deterministic-mixed), on par with 4B. The full reconciliation, including the methodology for the matched-path deterministic-mixed pre-SFT 0.8B number, is in Appendix~\ref{sec:appendix-path}.

\section{Cross-Family Judge Validation}\label{sec:cross-family}

The judge-overlap concern (Opus 4.7 used for skill rewriting, task synthesis, SFT-trace generation, and judging) is the strongest single-axis methodological threat in the experiment. To bound the magnitude, we ran a non-Anthropic-family second judge (GPT-5.4 via OpenRouter) on the LLM-only re-judge episodes for all seven configurations (Table~\ref{tab:cross-family}).

\begin{table}[h]
\centering
\small
\begin{tabular}{@{}l c c c c c@{}}
\toprule
& \multicolumn{2}{c}{\textbf{Opus 4.7 LLM-only}} & \multicolumn{2}{c}{\textbf{GPT-5.4 (OpenRouter)}} & \\
\cmidrule(lr){2-3}\cmidrule(lr){4-5}
& BL / CU & $\Delta$ & BL / CU & $\Delta$ & $\Delta$ shift \\
\midrule
v1   (0.8B+LoRA)        & 0.710 / 0.705 & $-0.005$ & 0.700 / 0.700 & $\;0.000$ & $+0.005$ \\
v1.9 (2B+LoRA)          & 0.815 / 0.915 & $+0.100$ & 0.815 / 0.915 & $+0.100$ & $\;0.000$ \\
v2.0 (4B+LoRA)          & 0.920 / 0.985 & $+0.065$ & 0.920 / 0.980 & $+0.060$ & $-0.005$ \\
Haiku-4-5 (ref)         & 0.955 / 0.985 & $+0.030$ & 0.955 / 0.975 & $+0.020$ & $-0.010$ \\
\textbf{pre-SFT 0.8B}   & \textbf{0.665 / 0.590} & $\boldsymbol{-0.075}$ & \textbf{0.670 / 0.585} & $\boldsymbol{-0.085}$ & $-0.010$ \\
pre-SFT 2B              & 0.755 / 0.815 & $+0.060$ & 0.750 / 0.835 & $+0.085$ & $+0.025$ \\
\textbf{pre-SFT 4B}     & \textbf{0.850 / 0.840} & $\boldsymbol{-0.010}$ & \textbf{0.850 / 0.805} & $\boldsymbol{-0.045}$ & $\boldsymbol{-0.035}$ \\
\bottomrule
\end{tabular}
\caption{Cross-family judge cross-validation across all seven configurations / 2800 paired episodes. Per-episode agreement ranges from $93.25\%$ (pre-SFT 4B) to $100.00\%$ (v1.9, 400/400 identical verdicts). Cohen's $\kappa$ ranges from $0.754$ to $1.000$. Pre-SFT 0.8B HF agreement is $98.50\%$ ($\kappa = 0.968$). Headline $\Delta$ shifts $\leq 0.035$\,pp on every dataset; both judges agree on the direction of every per-student finding, including the W-shape's two negative-Δ pockets at 0.8B and 4B.}
\label{tab:cross-family}
\end{table}

Across 2800 paired episodes, the maximum headline-Δ shift between judges is $0.035$\,pp (pre-SFT 4B); the minimum agreement is $93.25\%$; the minimum $\kappa$ is $0.754$. Under a non-Anthropic-family second judge on non-Anthropic infrastructure, no headline conclusion changes direction across any of the seven configurations. Both negative-Δ pockets in the W-shape are judge-invariant: pre-SFT 0.8B has $\Delta \leq -0.075$ (Opus $-0.075$, GPT $-0.085$), pre-SFT 4B has $\Delta \leq 0$ (Opus $-0.010$, GPT $-0.045$). Cross-family bounds on the SFT contribution: at 0.8B, Δ-lift $[+0.070, +0.085]$; at 2B, Δ-lift $[+0.015, +0.040]$ (GPT credits base 2B with more procedural responsiveness than Opus does); at 4B, Δ-lift $[+0.075, +0.105]$. All three SFT contributions are positive under both judges.

\paragraph{What this validation does not close.} It addresses only the \emph{judging-stage} overlap. Opus still wrote the procedural rewrites, synthesized the tasks, and generated the SFT-corpus traces; this corpus-generation overlap is structurally unbounded by judging-stage cross-validation. A non-Anthropic third-family judge (Gemini, Llama-3.3-70B) or human raters would extend the bound; we did not run them. The most plausible remaining direction of bias is in the corpus-generation stage rather than the judging stage.

\section{Recipe Iteration at 0.8B}\label{sec:0.8b-iteration}

Before scaling to larger bases, we ran five LoRA-recipe variants at 0.8B (v1--v1.8) to test whether the small-base bottleneck was in the corpus, the optimizer, or the chat-template handling. Table~\ref{tab:headline} reports all eleven configurations under the bench's native deterministic-mixed dispatch.

\begin{table}[h]
\centering
\small
\begin{tabularx}{\textwidth}{@{}l l Y r r r@{}}
\toprule
\textbf{Variant} & \textbf{Base} & \textbf{Recipe} & \textbf{BL} & \textbf{CU} & \textbf{$\Delta$} \\
\midrule
Pre-SFT 0.8B & 0.8B (HF)     & none & 0.625 & 0.510 & $-0.115$\,$\S$ \\
Pre-SFT 2B   & 2B (HF)       & none & 0.685 & 0.710 & $+0.025$ \\
Pre-SFT 4B   & 4B (HF)       & none & $0.000$\,$\dagger$ & $0.000$\,$\dagger$ & --- \\
Pre-SFT Haiku-4-5 (ref) & claude-haiku-4-5 & none & 0.785 & 0.800 & $+0.015$ \\
\addlinespace
\textbf{v1}    & 0.8B & LoRA $r{=}8$, curated-only, 353 rows           & 0.635 & 0.585 & $-0.050$ \\
\textbf{v1.5}  & 0.8B & LoRA + BL+CU mixed corpus                       & 0.650 & 0.425 & $-0.225$ \\
\textbf{v1.6}  & 0.8B & LoRA + drop ceiling skills + chat-patch         & 0.645 & 0.565 & $-0.080$ \\
\textbf{v1.7}  & 0.8B & partial FT (top-6 layers + lm\_head)            & 0.465 & 0.615 & $+0.150$\,$\ddagger$ \\
\textbf{v1.8}  & 0.8B & partial FT + skill-block stripped from training & 0.545 & 0.570 & $+0.025$ \\
\textbf{v1.9}  & 2B   & LoRA $r{=}16$, curated-only, 353 rows           & \textbf{0.750} & \textbf{0.825} & $\boldsymbol{+0.075}$ \\
\textbf{v2.0}  & 4B   & LoRA $r{=}32$, curated-only, 353 rows           & \textbf{0.835} & \textbf{0.880} & $\boldsymbol{+0.045}$ \\
\bottomrule
\end{tabularx}
\caption{Recipe-by-recipe pass rates on the 200-task holdout under deterministic-mixed scoring (the bench's native dispatch). All bases are Qwen3.5 dense models except where noted. All evaluations use HF transformers generation; the original Ollama-path pre-SFT 0.8B numbers ($0.510 / 0.565 / +0.055$) are documented in Appendix~\ref{sec:appendix-path}. $\dagger$ Pre-SFT 4B's $0/0$ is a format-compliance artifact (\S\ref{sec:format-artifact}). $\ddagger$ v1.7's apparent $+\Delta$ is an artifact of \BL{} collapse, not differential learning. $\S$ The matched-path methodology for pre-SFT 0.8B is in Appendix~\ref{sec:appendix-path}. LLM-only re-judge values (Table~\ref{tab:llm-only}) reframe the SFT attribution at 0.8B; Δ-lift is $+0.070$ matched-path LLM-only, $+0.065$ deterministic-mixed.}
\label{tab:headline}
\end{table}

\subsection{Recipe variation does not move the 0.8B ceiling}\label{sec:0.8b-recipe-variants}
Three well-formed 0.8B variants (v1, v1.6, v1.8) cluster \CU{} within a 2\,pp band ($0.565$--$0.585$ deterministic, $\sim 0.705$ for v1 under LLM-only re-judge). The cluster does not move with corpus composition (v1.5 adds \BL+\CU mixed rows; collapses \CU{} to $0.425$), ceiling-skill dropping (v1.6 removes skills where the base passes $\geq 0.90$; lands within v1's band), or skill-block stripping (v1.8 trains with the SKILL block removed from system prompts; lands within v1's band). The post-SFT absolute \CU{} pass rate at 0.8B is bounded by base 0.8B's reasoning capability, regardless of recipe.

\paragraph{v1.7's apparent Δ is a mode-collapse artifact.} Replacing LoRA with full-rank partial fine-tuning of the top-6 layers + \texttt{lm\_head} + final \texttt{norm} (using 8-bit Adam to fit consumer GPU budget) produces an apparent $\Delta = +0.150$, but with \BL{} \emph{collapsed} to $0.465$ --- substantially below the matched-path pre-SFT 0.8B \BL{} of $0.625$. Inspection of v1.7 \BL{} failures reveals the model emits ``\texttt{Step 1 (SKILL block: \dots)}'' prefaces and procedural step language even when no skill block is in the system prompt: the model has learned to expect a skill block and hallucinates one when absent. The $+0.150$ Δ does not reflect the model learning to use skill blocks; it reflects the model losing the ability to function without one. v1.8 (skill-block stripped from training) recovers \BL{} to $0.545$ and lands \CU{} back inside the $0.565$--$0.585$ band, confirming the diagnosis.

\paragraph{What the 0.8B negative results constrain, and what they do not.} What clusters at 0.8B is the post-SFT \emph{absolute} pass rate, which is base-capacity-bound. What does \emph{not} cluster is the SFT-attributable Δ-lift, which is $+0.070$ at 0.8B under matched-path scoring --- comparable to the 4B Δ-lift of $+0.075$ (Table~\ref{tab:sft-contribution}). The capacity-floor diagnosis was correct for absolute pass rate; under earlier path-mismatched scoring it was wrongly extended to the SFT mechanism. Under matched-path scoring, SFT delivers comparable differential work at 0.8B as at 4B; what it cannot do at 0.8B is push the absolute ceiling.

\subsection{v1.9 (2B) attribution split}\label{sec:v19-split}
Same recipe shape as v1 (LoRA, curated-only, 353 rows, 3 epochs), bumped to $r{=}16$/$\alpha{=}32$ to give the larger base more LoRA capacity, on \Qwent. Result: \BL{} $0.750$ / \CU{} $0.825$ / $\Delta$ $+0.075$ (paired-bootstrap 95\% CI $[+0.020, +0.130]$; McNemar $\chi^2{=}5.60$, $p{=}0.018$).

\begin{table}[h]
\centering
\small
\begin{tabular}{@{}l rr rr@{}}
\toprule
& \multicolumn{2}{c}{\textbf{deterministic-mixed}} & \multicolumn{2}{c}{\textbf{LLM-only (matched-path)}} \\
\cmidrule(lr){2-3} \cmidrule(lr){4-5}
& $\Delta$BL & $\Delta$CU & $\Delta$BL & $\Delta$CU \\
\midrule
Base scaling 0.8B $\to$ 2B  & $+0.060$ & $+0.200$ & $+0.090$ & $+0.225$ \\
SFT contribution at 2B      & $+0.065$ & $\mathbf{+0.115}$ & $+0.060$ & $\mathbf{+0.100}$ \\
\midrule
Total v1.9 vs pre-SFT 0.8B  & $+0.125$ & $+0.315$ & $+0.150$ & $+0.325$ \\
SFT $\Delta$-lift at 2B     & \multicolumn{2}{c}{$+0.025 \to +0.075$ ($\boldsymbol{+0.050}$)} & \multicolumn{2}{c}{$+0.060 \to +0.100$ ($\boldsymbol{+0.040}$)} \\
\bottomrule
\end{tabular}
\caption{v1.9 attribution split under both judging paths. The deterministic-mixed columns use matched-path pre-SFT 0.8B (HF, $0.625/0.510$; Appendix~\ref{sec:appendix-path}) and pre-SFT 2B (HF, $0.685/0.710$); the LLM-only columns use the corresponding LLM-only-judged controls. The SFT-attributable Δ-lift at 2B is $+0.050$ under deterministic-mixed and $+0.040$ under LLM-only --- consistent within 1\,pp across both judging paths even as absolute magnitudes shift.}
\label{tab:attribution}
\end{table}

\begin{figure}[h]
\centering
\includegraphics[width=0.5\linewidth]{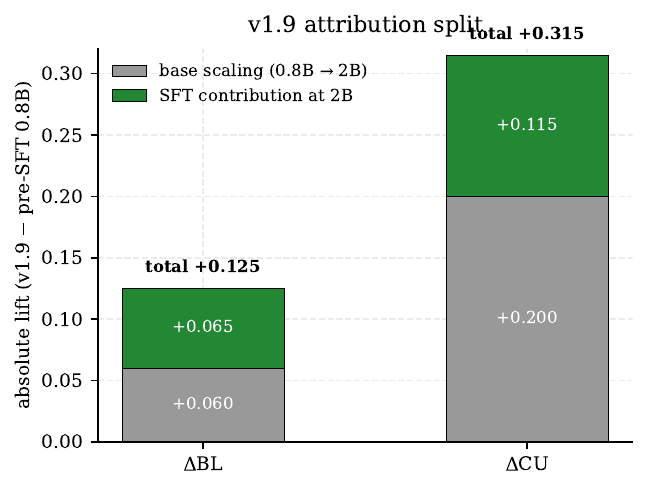}
\caption{v1.9 lift over pre-SFT 0.8B decomposed into base-model scaling (gray; measured between the two pre-SFT controls) and SFT contribution at 2B (green; the residual from pre-SFT 2B to v1.9). The SFT contribution is $+6.5$\,pp on \BL{} and $+11.5$\,pp on \CU.}
\label{fig:attribution}
\end{figure}

\section{Statistical Tests}\label{sec:stats}

Each task is run in both \BL{} and \CU{} conditions, so verdicts are paired. We run McNemar's test (continuity-corrected $\chi^2$ for $n_{\mathrm{discordant}} \geq 25$, exact binomial otherwise) per model, and a paired bootstrap (10{,}000 resamples of \texttt{task\_uid} with replacement) for 95\% CIs on Δ. For the cross-model saturation test we resample v1.9 and v2.0 independently and report a one-sided bootstrap test for $H_0: \Delta_{\mathrm{v1.9}} \leq \Delta_{\mathrm{v2.0}}$.

\begin{table}[h]
\centering
\small
\begin{tabular}{@{}l c c c c@{}}
\toprule
& \multicolumn{2}{c}{\textbf{deterministic-mixed}} & \multicolumn{2}{c}{\textbf{LLM-only re-judge}} \\
\cmidrule(lr){2-3}\cmidrule(lr){4-5}
& $\Delta$ & McNemar $p$ / 95\% CI & $\Delta$ & McNemar $p$ / 95\% CI \\
\midrule
v1.9 & $+0.075$ & $\chi^2{=}5.60$, $p{=}0.018$ & $+0.100$ & $\chi^2{=}9.03$, $p{=}0.0027$ \\
     &          & CI $[+0.020, +0.130]$        &          & CI $[+0.040, +0.160]$ \\
v2.0 & $+0.045$ & exact, $p{=}0.049$           & $+0.065$ & exact, $p{=}0.00098$ \\
     &          & CI $[+0.005, +0.085]$        &          & CI $[+0.030, +0.105]$ \\
\bottomrule
\end{tabular}
\caption{Per-model paired tests on the procedural-Δ, under both judging paths. Both models have a positive Δ with bootstrap CI excluding zero in both paths. Under LLM-only re-judge, v2.0's $p{=}0.00098$ comfortably survives Bonferroni correction across the two tests (the deterministic-mixed $p{=}0.049$ would not). Neither test is replicated across seeds.}
\label{tab:stats-per-model}
\end{table}

For the cross-model saturation test, the result is essentially unchanged across judging paths: deterministic-mixed $\Delta_{\mathrm{v1.9}} - \Delta_{\mathrm{v2.0}} = +0.030$, CI $[-0.040, +0.100]$, one-sided $p = 0.225$; LLM-only $\Delta_{\mathrm{v1.9}} - \Delta_{\mathrm{v2.0}} = +0.035$, CI $[-0.040, +0.105]$, one-sided $p = 0.191$. The CI includes zero in both paths. The Δ-shrinkage v1.9$\to$v2.0 is consistent with bench saturation in direction but is not statistically distinguishable from sampling noise at $n{=}200$. The cross-scale SFT contribution growth (Δ-lift $+0.040 \to +0.075$ from 2B to 4B; \S\ref{sec:cross-scale}) is robust to the same noise because it is computed against a different pre-SFT baseline (pre-SFT 4B Δ $-0.010$).

\begin{figure}[h]
\centering
\includegraphics[width=\linewidth]{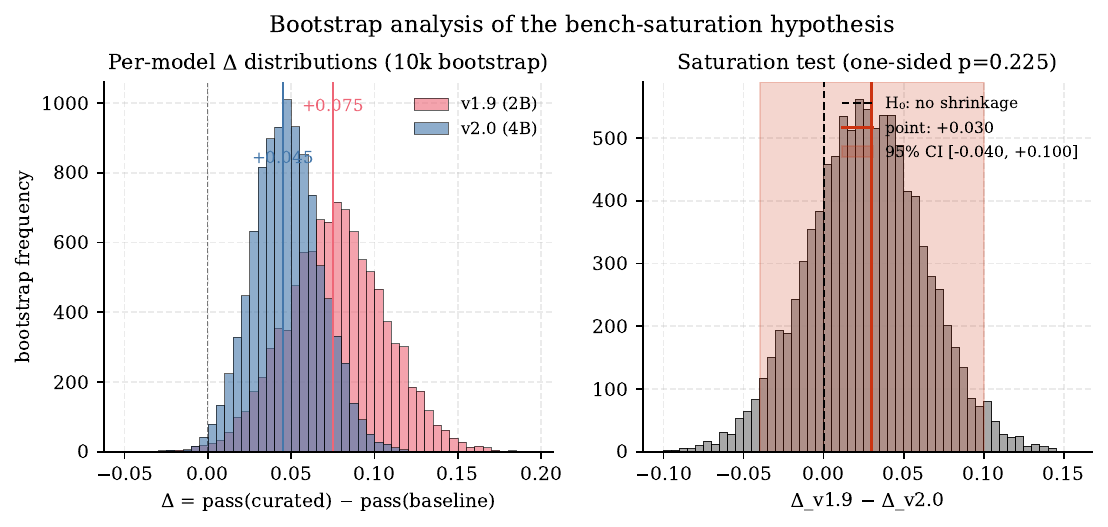}
\caption{Left: bootstrap distributions of the procedural-Δ for v1.9 and v2.0 (10{,}000 resamples of paired \texttt{task\_uid}s). Both distributions sit above zero (dashed line) but overlap substantially. Right: distribution of $\Delta_{\mathrm{v1.9}} - \Delta_{\mathrm{v2.0}}$; the 95\% CI (red shading) includes zero, and the one-sided test for the saturation hypothesis returns $p{=}0.225$. Direction consistent with saturation; magnitude not statistically distinguishable from noise.}
\label{fig:bootstrap}
\end{figure}

\section{Generality Probe}\label{sec:generality}

We probe v1.9, v2.0, base 2B, and base 4B with a 52-prompt OOD battery (10 generality + 42 factual including 8 false-premise ``trick'' prompts) under a generic system prompt (not the \textsc{solver} prompt; Table~\ref{tab:generality}). The probe is a convenience sample, not a calibrated benchmark; we treat it as qualitative evidence of OOD behavior.

\begin{table}[h]
\centering
\small
\begin{tabular}{@{}lcccc@{}}
\toprule
& format-locked & factual HIT & trick CONFAB (eyeballed) & phantom skill block \\
\midrule
v1.9 (2B) & $0/52$ & $33/34$\,$\dagger$ & $6/8$ & $0/52$ \\
base 4B   & $0/52$ & $34/34$ & $\sim\!1/8$ & $0/52$ \\
v2.0 (4B) & $0/52$ & $34/34$ & $\mathbf{0/8}$ & $0/52$ \\
\bottomrule
\end{tabular}
\caption{Generality probe under generic system prompt. $\dagger$ v1.9 swapped \emph{Brave New World} attribution from Aldous Huxley to H.\,G.\,Wells --- a plausible-author confabulation absent from base 4B and v2.0.}
\label{tab:generality}
\end{table}

v1.9 introduced a small calibration regression on adversarial prompts: the SFT-induced ``produce a confident step-by-step answer'' prior leaks into out-of-distribution behavior, making v1.9 commit to plausible-sounding inventions where base 2B would refuse. v2.0 does not exhibit this --- the 4B base has enough native capacity to recognize false premises and refuse cleanly without the SFT shape overriding refusal. The bench-time finding that 17.5\% of v2.0's \BL{} responses reference a ``skill block'' not present in the system prompt does \emph{not} extend to general-chat distributions: under a generic system prompt, the SFT-induced skill-block reflex is dormant ($0/52$). The phantom-mention is contextual to SFT-distribution-similar prompts.

\section{Discussion}

\subsection{Mode collapse reinterpreted}
The v1.7 mode collapse (partial-FT producing apparent $+\Delta$ at the cost of \BL{} integrity; \S\ref{sec:0.8b-iteration}) appeared at 0.8B as a structural concern about SFT-on-procedural-demonstrations and motivated the v1.8 skill-block-stripping intervention. Spot-checks of v2.0 \BL{} responses show the same phantom skill-block reference ($35/200$, 17.5\%) but with $94.3\%$ of those responses still passing: at 4B, the phantom mention is cosmetic rather than load-bearing. The model has the capacity to execute the procedure regardless of whether it remembers seeing a skill block in the system prompt. The mode collapse was a base-capacity-floor artifact at 0.8B, not a structural property of the SFT corpus. We do not claim the v1.8 skill-strip fix is unnecessary in general; we observe only that, at the model size where SFT produces a measurable differential lift in this study (4B), the failure mode it was designed to prevent is not load-bearing.

\subsection{Negative results were informative}
The five 0.8B variants are individually informative as negative results: corpus composition, partial fine-tuning, and chat-template patching all failed to lift the absolute pass rate at this scale and corpus size. They are jointly informative: together they constrain the diagnosis to ``base-model capacity is the binding constraint on absolute pass rate at this corpus size,'' which the 2B run validates. Running through the negative-result series before scaling was epistemically efficient ($\sim$\$15 in compute, three days of GPU time) compared to skipping straight to 4B with no priors on what fails. We argue iterated negative-result reporting is undervalued in the SFT-on-demonstrations literature: most claims in this space are backed by single-recipe positive results without surrounding falsification of nearby variants.

\subsection{What we did not separately decompose}
The differential Δ-lift is evidence that there is a procedure-vs-no-procedure component of the gain that pure format-learning cannot explain. We do not attempt a quantitative ``$x\%$ format, $y\%$ procedure'' decomposition: the format and procedure components are not independent (producing the procedural-step shape implicitly enforces a checking discipline that may itself be the procedural lift), and per-condition pass-rates do not separate them cleanly. A natural further ablation --- training the same Qwen3.5 base on 353 rows of generic Opus instruction-tuning data and re-running the bench --- would sharpen the specificity claim; we did not run it (\S\ref{sec:threats}).

\section{Threats to Validity and Limitations}\label{sec:threats}

\begin{enumerate}[leftmargin=*]
  \item \textbf{Single-seed evaluation.} All headline numbers (Tables~\ref{tab:llm-only}, \ref{tab:sft-contribution}, \ref{tab:stats-per-model}) are single-seed point estimates with within-seed paired statistics. The v2.0 deterministic McNemar $p{=}0.049$ should be read as ``one trial that just barely cleared the conventional threshold,'' not as established significance; the LLM-only $p{=}0.00098$ is more robust but is also a single seed. 3--5 seed replication of v1.9 and v2.0 would bound run-to-run variance \citep{biderman2024lessons}; we did not run it.

  \item \textbf{Corpus-generation overlap with the judge.} Opus 4.7 was used in four roles: rewriting the declarative skill catalog into procedural form, synthesizing the held-out tasks, generating the SFT-corpus traces, and judging both pre-SFT and post-SFT responses. The cross-family validation (\S\ref{sec:cross-family}, 2800 paired episodes, max headline Δ shift $\leq 0.035$\,pp, $\kappa \geq 0.754$) bounds the \emph{judging-stage} component. The corpus-generation overlap is structurally unbounded by judging-stage cross-validation. A non-Anthropic third-family judge (Gemini, Llama-3.3-70B) or human raters would extend the bound to two non-Anthropic families; estimated cost $\sim$\$5--10 (LLM judge) or $\sim$\$50--100 (human raters) on a 200-episode subsample. The most plausible remaining direction of bias is in the corpus-generation stage rather than the judging stage.

  \item \textbf{Single-skill-per-task synthesis.} Tasks were generated by Opus from a known skill, then the same skill was injected at evaluation time. The bench measures \emph{procedural application on procedure-aligned tasks}, not skill-routing or compositional skill use. This is closer to the Skill-Mix compositional hypothesis \citep{yu2024skillmix} than to a tautology --- the student does have to follow the procedure correctly, and many tasks include constructed traps for procedure-skipping models --- but it is narrower than ``procedural skill use'' as it appears in some recent work. The honest claim is that SFT'd students apply named procedures more effectively when shown a procedure aligned to a procedure-aligned task.

  \item \textbf{Single model family (Qwen3.5).} All three SFT'd students and the two HF-evaluated pre-SFT controls are Qwen3.5 dense models. The W-shape we observe in pre-SFT Δ may be a Qwen3.5-specific phenomenon or a property of small-to-mid-scale dense LLMs in general; this paper cannot distinguish. A cross-architecture replication at a single capacity tier (e.g.\ Llama-3.2-3B as a 4B-class comparison) would be cheap and would substantially strengthen the architectural-generality claim. The Haiku-4-5 reference contributes one architecturally distinct data point but is not size-matched.

  \item \textbf{No generic instruction-tuning control.} The paper's central distinction is between ``the model uses the injected procedure'' and ``the model just imitates the response shape.'' Differential Δ-lift is the load-bearing evidence for the first reading. The specificity claim --- that \emph{procedural-skill} SFT delivers the $+0.04$ to $+0.075$ Δ-lift, as opposed to any 353-row Opus-trace SFT --- is unwitnessed by the most natural ablation: training the same Qwen3.5 base on 353 rows of generic Opus instruction-tuning data and re-running the bench. Three outcomes the present data does not rule out: (a) generic SFT gives the same Δ-lift (the recipe teaches CoT discipline, not procedural specificity); (b) generic SFT gives smaller or negative Δ-lift (procedural specificity is real); (c) generic SFT gives a different cross-scale trajectory. Estimated cost: one additional training run per scale, $\sim$1 GPU-day total.

  \item \textbf{$n{=}200$ per condition limits Δ-statistic resolution.} Per-condition binomial standard error is $\sim$3\,pp. The cross-model bootstrap test for the v1.9-vs-v2.0 saturation hypothesis fails ($p{=}0.191$ LLM-only, $p{=}0.225$ deterministic). This only addresses the post-SFT Δ-shrinkage; the SFT-contribution comparisons across scales are robust to the same noise because they are computed from distinct pre-SFT baselines.

  \item \textbf{Decoding is greedy throughout.} We did not sweep temperature or top-$p$; whether the post-SFT Δ is robust to sampling stochasticity is untested.

  \item \textbf{Generality probe is a convenience sample.} The 52-prompt OOD battery (Table~\ref{tab:generality}) is hand-written; trick-prompt CONFAB judgments were eyeballed. The qualitative claim ``v2.0 preserves base 4B's out-of-distribution behavior'' is supported by the data but should not be read as quantitative parity.

  \item \textbf{Haiku inference path mismatch (minor).} Haiku-4-5 was evaluated through the Anthropic API while Qwen3.5 students used HF transformers locally. Inference parameters were matched as closely as possible; tokenization and decoding stack still differ. The load-bearing comparisons in this paper are within-Qwen3.5 across capacity tier, not Qwen3.5-vs-Haiku.
\end{enumerate}

\paragraph{What this paper does not claim.} We do not claim that pre-SFT 4B is below Haiku on general capability; only that, on this bench with this corpus, v2.0 (4B+SFT) and Haiku tie at the curated ceiling. We do not claim the W-shape generalises to architectures we did not test; we offer a falsifiable prediction at 8B/14B and at other architectures. We do not claim a quantitative format-vs-procedure decomposition; only that the differential Δ-lift evidences a procedure-vs-no-procedure component pure format-learning cannot explain.

\section{Conclusion}

Across three Qwen3.5 dense scales (0.8B, 2B, 4B) under matched-path scoring, procedural-skill SFT on a 353-row corpus delivers a roughly uniform Δ-lift in the +0.040 to +0.075 band. Variation in post-SFT Δ across sizes is dominated by a W-shaped pre-SFT trajectory: the procedure hurts very small bases (compounding errors) and competent-but-not-frontier bases (overhead without signal), helps an intermediate base (scaffolding skipped checks), and helps a frontier base modestly (native procedural capacity). SFT's distinctive value is in lifting Δ in the negative regimes --- rescuing 0.8B from procedure-incompetent to procedure-neutral, flipping 4B from procedure-overhead to procedure-helps. The mechanism is regime-asymmetric rather than capacity-conditional, with a falsifiable prediction at 8B/14B. v2.0 (4B+SFT) ties the Haiku-4-5 frontier reference at the bench's effective ceiling under fair scoring.

The methodological observations alongside the empirical result are: a bench format-compliance artifact biases SFT-lift claims absent a format-tolerant base-model control; the artifact's remediation (LLM-only re-judge) reveals the deterministic judge had been systematically biased \emph{against} the curated condition by 2--4.5\,pp; cross-family judge cross-validation (GPT-5.4, 2800 paired episodes) bounds the judging-stage component of the corpus-author-overlap concern. Two earlier framings of this work --- ``format-only learning at 0.8B'' (a deterministic-vs-Ollama path mismatch; Appendix~\ref{sec:appendix-path}) and ``SFT contribution shrinks at 4B'' (the pre-SFT 4B control had not yet been measured) --- were path-mismatch artifacts and are explicitly superseded. The remaining structurally unaddressed concern is corpus-generation overlap; future work should extend the cross-family bound to a third family and replicate across seeds, both inexpensive.

\section*{Acknowledgments}
Compute provided by vast.ai (\Qwent{} and \Qwenf{} training and partial evaluation) and a local consumer GPU (RTX 3080\,Ti, evaluation and probing). Skill catalog content seeded from Skill-Mix \citep{yu2024skillmix} Tables 5 and 6 and expanded by hand. Opus 4.7 (1M context) was used for skill procedural rewriting, task synthesis, baseline trace generation, and judge-of-record scoring; the resulting judge-overlap is itemised as a threat to validity in \S\ref{sec:threats}.

\section*{Code and Data Availability}
Repository: \url{https://github.com/izzortsi/skillmix-n-skillsbench} (branch \texttt{dev}). Episode-level eval logs at \texttt{data/pipeline-runs/default/bench-eval-*}. Engineering reports in \texttt{b1.reports/} (cited in Appendix~\ref{app:reports}); dated, self-contained snapshots written contemporaneously with each result, included for reproducibility audit rather than as primary evidence.


\appendix

\section{Path-mismatch resolution for pre-SFT 0.8B}\label{sec:appendix-path}

The original pre-SFT 0.8B baseline (\BL{} $0.510$ / \CU{} $0.565$ / $\Delta +0.055$) was run via Ollama, not the same HuggingFace transformers path used for 2B/4B. Quantization defaults, chat-template handling, and decoding-stack details differ between the two paths. The mismatch was material: under matched HF scoring, pre-SFT 0.8B lands at \BL{} $0.625$ / \CU{} $0.510$ / $\Delta -0.115$ (deterministic-mixed) or \BL{} $0.665$ / \CU{} $0.590$ / $\Delta -0.075$ (LLM-only). The Ollama-path $\Delta +0.055$ sat on the wrong side of zero relative to both matched-path measurements, and the v1 attribution flipped accordingly.

\subsection{Methodology}
We resolved the mismatch in two parts:

\paragraph{LLM-only path.} A fresh HF eval of pre-SFT 0.8B under \texttt{--force-llm-judge} (Table~\ref{tab:llm-only}) lands at $0.665 / 0.590 / -0.075$.

\paragraph{Deterministic-mixed path (Table~\ref{tab:headline}'s native dispatch).} The 334 deterministic-type episodes from the same HF run were re-scored locally via the deterministic extractor; the 33 \texttt{FREE\_FORM} tasks (66 episodes) carry their LLM-judged scores from the \texttt{--force-llm-judge} run, since under \texttt{--force-llm-judge} \texttt{FREE\_FORM} tasks land on the same LLM/FREE\_FORM verifier path that Table~\ref{tab:headline}'s deterministic-mixed dispatch uses for \texttt{FREE\_FORM}. Result: $0.625 / 0.510 / -0.115$. Reproducer: \texttt{training/rejudge\_failed\_episodes.py --judge deterministic --all}. This is Table~\ref{tab:headline}'s pre-SFT 0.8B row; the original Ollama numbers are no longer presented in any headline table.

\subsection{Consequence for v1 attribution}
Under matched HF scoring, the v1 SFT Δ-lift is $+0.065$ (deterministic-mixed) or $+0.070$ (LLM-only) --- consistent across both judging paths within 1\,pp and not the $-0.060$ ``Δ-flipped negative'' the path-mismatched comparison originally suggested. The earlier ``format-only learning at 0.8B'' framing compared v1's deterministic $\Delta = -0.050$ against Ollama-path pre-SFT $\Delta = +0.055$, both deterministic-mixed but on different generation paths. Under the corrected matched-path pre-SFT row, SFT delivers $+1.0$\,pp \BL{} lift, $+7.5$\,pp \CU{} lift, and $+0.065$ Δ-lift attributable to SFT at 0.8B --- a clean differential lift rather than a Δ-flip. The 2B and 4B SFT contributions are unaffected since their endpoints were already on a single path each.

\section{Pipeline file structure}\label{app:pipeline}
{\small
\begin{verbatim}
skillmix-n-skillsbench/
|-- core/                                shared schema + provider abstractions
|-- harness/                             LinearAgent: streaming + thinking-block capture
|-- b2_benchmarks/skillsbench/           corpus_harness, llm_judge, procedural_catalog
|-- s1_extracting_skill_names/           declarative-skill seeders (Wikipedia-style)
|-- s2_extracting_skills_from_text/      procedural-rewrite stage
|-- s3_generating_skill_examples/m5_*.py task synthesis from procedural skills
|-- s4_extracting_skill_usage_instances/ solver + trace + judge composition (m4)
|-- training/
|   |-- train_qwen35_lora.py             LoRA / partial-FT trainer (TRL >=0.18)
|   |-- eval_qwen35_lora.py              holdout evaluator (--adapter optional)
|   |-- patch_qwen35_template.py         chat-template generation-marker patcher
|   |-- chat_with_adapter.py             interactive + probe-set generality tester
|   |-- rejudge_failed_episodes.py       post-hoc rejudge for transient judge errors
|   `-- compare_pre_post.py              per-skill diff utility
|-- docker/                              vast.ai training image + setup docs
|-- data/pipeline-runs/default/          full run artifacts
`-- b1.reports/                          engineering log (yymmdd.title.txt)
\end{verbatim}
}

\section{Per-skill v2.0 result ($n{=}5$ per skill, sorted by $\Delta$)}\label{app:per-skill}
Lift cluster (procedure helped, deterministic judging):
\begin{itemize}[leftmargin=*,nosep]
  \item accident-fallacy $+0.40$, spatial-reasoning $+0.40$
  \item false-dilemma, confirmation-bias, anchoring-bias, appeal-to-emotion, base-rate-neglect, transitive-inference, necessary-and-sufficient-conditions, metaphor, emotional-self-regulation: $+0.20$ each
\end{itemize}
Flat cluster ($\Delta = 0$): 25 skills, of which 22 are at \BL{} $\geq 0.80$ (ceiling-bound).

Regression cluster (procedure hurt; all $5/5$ \BL{} $\to 4/5$ \CU): complex-question, hypothetical-syllogism, framing-effect, equivocation. The regression cluster overlaps with deductive-logic skills where the base 4B has ceiling competence; this matches the per-skill miniature of the cross-scale regime-asymmetric finding (\S\ref{sec:per-skill}). Under LLM-only judging the regression cluster shrinks to one skill (equivocation), consistent with the curated-condition deterministic-judge bias (\S\ref{sec:format-artifact}).

\section{Engineering reports cited}\label{app:reports}
The repository's \texttt{b1.reports/} directory contains dated, self-contained engineering snapshots written contemporaneously with each result. These are included for reproducibility audit rather than as primary evidence; the primary evidence in this paper is the episode-level logs and the recipe-by-recipe pass rates in Tables~\ref{tab:llm-only} and \ref{tab:headline}.
\begin{itemize}[leftmargin=*,nosep]
  \item \texttt{260426.findings-pre-post-sft-iterations.txt} --- v1--v1.8 consolidation
  \item \texttt{260428.sft-v1\_9-2b-result.txt} --- v1.9 result + attribution
  \item \texttt{260428.v1\_9-generality-probe.txt} --- v1.9 generality regression
  \item \texttt{260428.sft-v2\_0-4b-result.txt} --- v2.0 result + attribution caveat
  \item \texttt{260428.v2\_0-generality-probe.txt} --- v2.0 generality preservation
  \item \texttt{260506.bench-format-sensitivity-finding.txt} --- pre-SFT 4B format artifact
  \item \texttt{260506.bench-statistical-significance.txt} --- McNemar + bootstrap tests under deterministic judging
  \item \texttt{260506.llm-only-rejudge-findings.txt} --- LLM-only re-judge across all seven configurations; the pre-SFT 4B finding ($\Delta \leq 0$), the pre-SFT 0.8B HF finding ($\Delta -0.075$, deepest trough), and the mechanism-uniformity reframe all originate here
  \item \texttt{260508.cross-family-judge-validation.txt} --- GPT-5.4 vs Opus 4.7 cross-family validation, 7 datasets / 2800 paired episodes, headline numbers in Table~\ref{tab:cross-family}
\end{itemize}

\end{document}